\documentclass{article}

\usepackage{microtype}
\usepackage{graphicx}
\usepackage{subcaption}
\usepackage{booktabs} 

\usepackage{hyperref}

\usepackage[preprint]{icml2026}

\usepackage{amsmath}
\usepackage{amssymb}
\usepackage{mathtools}
\usepackage{amsthm}
\usepackage[dvipsnames]{xcolor}
\usepackage{multirow}
\usepackage{colortbl}  
\usepackage{xcolor}
\usepackage{array}   
\usepackage{booktabs}
\usepackage{indentfirst}
\usepackage{graphicx}
\usepackage{blindtext}
\usepackage{cuted}
\usepackage{caption}
\usepackage{subcaption}
\usepackage{stfloats}
\usepackage[accsupp]{axessibility}
\usepackage{makecell}
\usepackage{float}

\usepackage[capitalize,noabbrev]{cleveref}

\definecolor{bestBase}{HTML}{D44478}
\definecolor{secondBase}{HTML}{FFAA33}
\definecolor{thirdBase}{HTML}{FFDD33}

\colorlet{bestLight}{bestBase!20!white}
\colorlet{secondLight}{secondBase!20!white}
\colorlet{thirdLight}{thirdBase!20!white}

\theoremstyle{plain}

\theoremstyle{definition}

\theoremstyle{remark}

\usepackage[textsize=tiny]{todonotes}

\icmltitlerunning{Time2General}

\begin{document}

\twocolumn[
  \icmltitle{Time2General: Learning Spatiotemporal Invariant Representations for \\ Domain-Generalization Video Semantic Segmentation}



  \icmlsetsymbol{equal}{*}
  \icmlsetsymbol{correspond}{†}

  \begin{icmlauthorlist}
    \icmlauthor{Siyu Chen}{jimei}
    \icmlauthor{Ting Han}{zhongshan,correspond}
    \icmlauthor{Haoling Huang}{zhongshan2,equal}
    \icmlauthor{Chaolei Wang}{zhongshan,equal}
    \icmlauthor{Chengzheng Fu}{nanhang,equal}
    \icmlauthor{Duxin Zhu}{jimei}\\
    \icmlauthor{Guorong Cai}{jimei}
    \icmlauthor{Jinhe Su}{jimei,correspond}
  \end{icmlauthorlist}

  \icmlaffiliation{jimei}{School of Computer Engineering, Jimei University, Xiamen, China}
  \icmlaffiliation{zhongshan}{School of Geospatial Engineering and Science, Sun Yat-sen University, Zhuhai, China}
  \icmlaffiliation{zhongshan2}{School of System Science and Engineering, Sun Yat-Sen University, Guangzhou, China}
  \icmlaffiliation{nanhang}{College of Artificial Intelligence, Nanjing University of Aeronautics and Astronautics, Nanjing, China}

  \icmlcorrespondingauthor{Jinhe Su}{sujh@jmu.edu.cn}
  \icmlcorrespondingauthor{Ting Han}{ting.devin.han@gmail.com}

  \icmlkeywords{Domain-Generalization, Video Semantic Segmentation, Invariant Representations}
  
  \vskip -0.5in

]



\printAffiliationsAndNotice{}  

\begin{strip}
    \centering
    \includegraphics[width=0.95\textwidth]{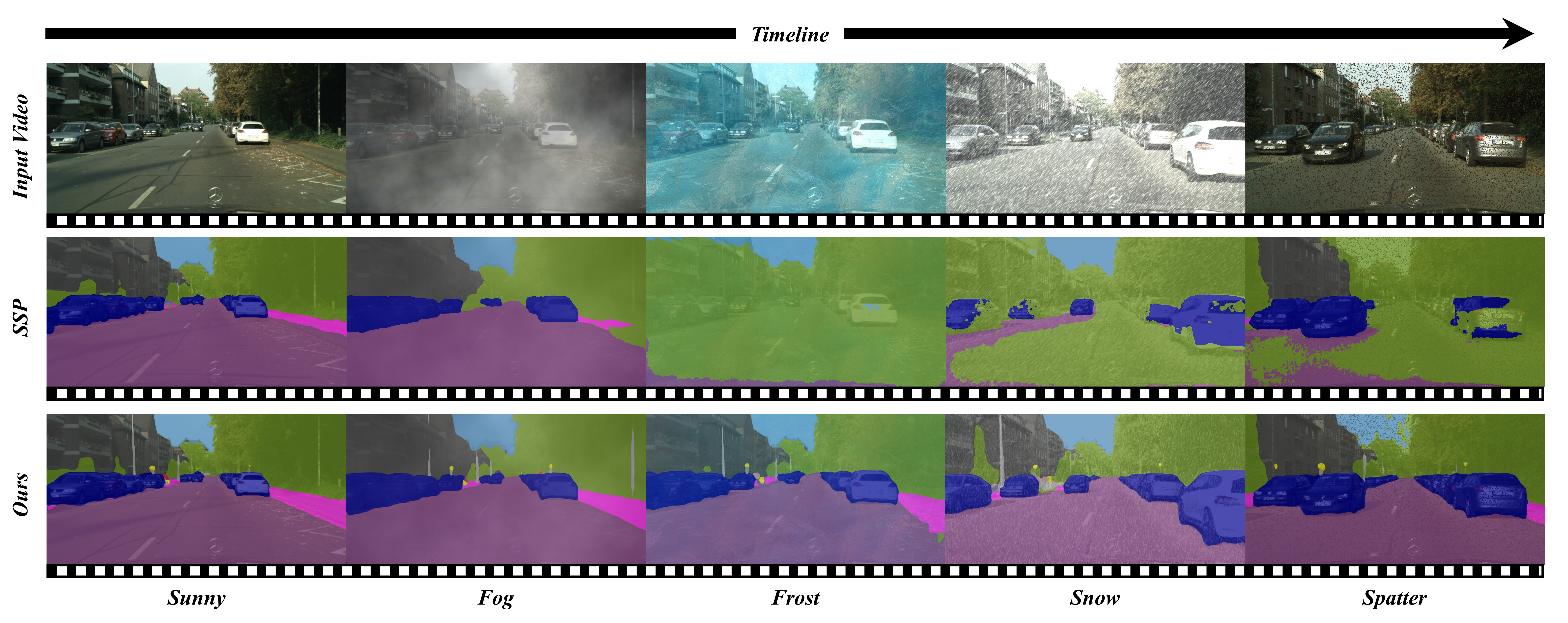}
    \vspace{-3mm}
    \captionof{figure}{We visualize semantic segmentation results on consecutive frames under multiple weather domains. Existing methods show a clear drop in accuracy in unseen weather, with drifting object boundaries, label switching, and poor temporal coherence. In contrast, our method produces more consistent predictions across time under domain shift.}
    \label{fig:motivation}
\end{strip}

\begin{abstract}
Domain Generalized Video Semantic Segmentation (DGVSS) is trained on a single labeled driving domain and is directly deployed on unseen domains without target labels and test-time adaptation while maintaining temporally consistent predictions over video streams. In practice, both domain shift and temporal-sampling shift break correspondence-based propagation and fixed-stride temporal aggregation, causing severe frame-to-frame flicker even in label-stable regions. We propose \textbf{Time2General}, a DGVSS framework built on \textbf{Stability Queries}. Time2General introduces a \textbf{Spatio-Temporal Memory Decoder} that aggregates multi-frame context into a clip-level spatio-temporal memory and decodes temporally consistent per-frame masks without explicit correspondence propagation. To further suppress flicker and improve robustness to varying sampling rates, the \textbf{Masked Temporal Consistency Loss} is proposed to regularize temporal prediction discrepancies across different strides, and randomize training strides to expose the model to diverse temporal gaps. Extensive experiments on multiple driving benchmarks show that Time2General achieves a substantial improvement in cross-domain accuracy and temporal stability over prior DGSS and VSS baselines while running at up to 18 FPS. Code will be released after the review process.
\end{abstract}

\section{Introduction}
Domain Generalized Video Semantic Segmentation (DGVSS) aims to train a video segmentation model from a single labeled source domain and generalize it to unseen target domains without any target-domain labels or test-time adaptation, while maintaining temporally consistent predictions over video streams. This task is motivated by real-world applications such as autonomous driving \cite{cordts2016cityscapes}, robotics, and mobile mapping, where appearance can vary substantially across cities, seasons, sensors, and adverse weather \cite{sakaridis2021acdc}.

Existing approaches for temporal stability in video semantic segmentation mainly adopt either propagation-based pipelines or clip-level spatiotemporal modeling. Propagation-based methods rely on inter-frame correspondences to warp/propagate features or logits \cite{zhu2017deep,gadde2017semantic,liu2020efficient}, making them sensitive to correspondence errors and prone to accumulation over time. Clip-level methods instead fuse multi-frame evidence via 3D operators or temporal attention  \cite{wang2018non,miao2021vspw,wang2021temporal}, but are constrained by computation when extending temporal context. In the domain-general setting, both directions are challenged: domain shift and visibility degradation can corrupt correspondence estimation, and spatiotemporal fusion learned on the source domain may not transfer, leading to unstable predictions in unseen domains (as snow in Figure \ref{fig:motivation}). Therefore, DGVSS calls for a correspondence-free and domain-robust mechanism to aggregate temporal context without relying on brittle matching \cite{sakaridis2021acdc,choi2021robustnet}.

DGVSS additionally faces temporal-sampling shift: videos collected across domains can differ substantially in acquisition frame rates (e.g., sub-Hz sampling versus high-FPS capture) \cite{cordts2016cityscapes,miao2021vspw}. Under such shift, designs that assume consecutive frames or use a fixed temporal stride correspond to very different physical time intervals across domains, which can induce substantially larger apparent motion and occlusion between sampled frames \cite{ilg2017flownet}. This breaks the implicit assumption that a fixed stride implies a comparable motion regime across domains, and consequently temporal aggregation modules learned under the source sampling pattern may become unreliable when applied to unseen sampling rates, resulting in degraded temporal stability.

In this paper, we aim to produce temporally stable and high-fidelity semantic masks for video streams. We propose \textbf{Time2General}, a domain-generalized video semantic segmentation approach. Specifically, to preserve the cross-domain generalization of foundation representations under the single-source supervision, we freeze a DINOv2 backbone \cite{oquab2023dinov2} to avoid overfitting caused by end-to-end fine-tuning. Based on frozen features, we introduce a set of learnable \textbf{Stability Queries} as a lightweight interface that absorbs complementary generalization cues from structural visual representations, geometric priors \cite{yang2024depth}, and text-aligned semantics \cite{radford2021learning,rao2022denseclip} into a unified query space. In this way, we learn consistent semantic grouping that is more generalized across domains and adverse conditions \cite{cheng2022masked}.

Built upon these query-conditioned multi-scale representations, we design a \textbf{Spatio-Temporal Memory Decoder} for video streams. Given a clip of $T$ frames, we inject learnable temporal embeddings into per-frame multi-scale tokens and concatenate them across time to construct a joint spatio-temporal memory. The same Stability Queries then attend to this memory and decode per-frame masks, enabling multi-frame context modeling without explicit correspondence estimation \cite{huang2022minvis,cheng2022masked}. At inference time, we adopt a sequential clip inference for long videos. To further suppress frame-to-frame flicker and address temporal-sampling shift\cite{lai2018learning}, we propose a \textbf{Masked Temporal Consistency Loss} that penalizes abrupt prediction changes on label-stable regions using multi-stride temporal differences with robust trimming, and we construct training clips by randomly sampling frames with varying temporal strides to expose the model to diverse temporal intervals.

We evaluate single-source cross-domain generalization on multiple driving video benchmarks under a unified label-alignment protocol \cite{lambert2020mseg}. We report results with three representative sources (ApolloScape\cite{huang2018apolloscape}, KITTI-360 \cite{liao2022kitti}, and CamVid \cite{brostow2009semantic}), and directly evaluate on the remaining datasets as unseen targets (including Cityscapes and Cityscapes-C \cite{michaelis2019benchmarking}) \emph{without} target-domain labels or test-time adaptation. Across diverse source$\rightarrow$target transfers, Time2General achieves consistent gains over prior image-based and video-based baselines, and runs in real time at \textbf{18 FPS} on an  NVIDIA RTX PRO A6000 with $1024\times512$ inputs. Our main contributions are summarized as follows:
\begin{itemize}
    \item We propose \textbf{Time2General} for domain-generalized video semantic segmentation, built on \textbf{Stability Queries} as temporally persistent semantic anchors.
    \item We introduce a \textbf{Spatio-Temporal Memory Decoder} that aggregates clip-level spatio-temporal context without explicit correspondence, enabling stable and efficient inference on long videos.
    \item We propose a \textbf{Masked Temporal Consistency Loss} and randomized temporal-stride sampling to reduce flicker under temporal-sampling shifts.
\end{itemize}

\begin{figure*}[!t]
  \centering
  \includegraphics[width=0.95\textwidth]{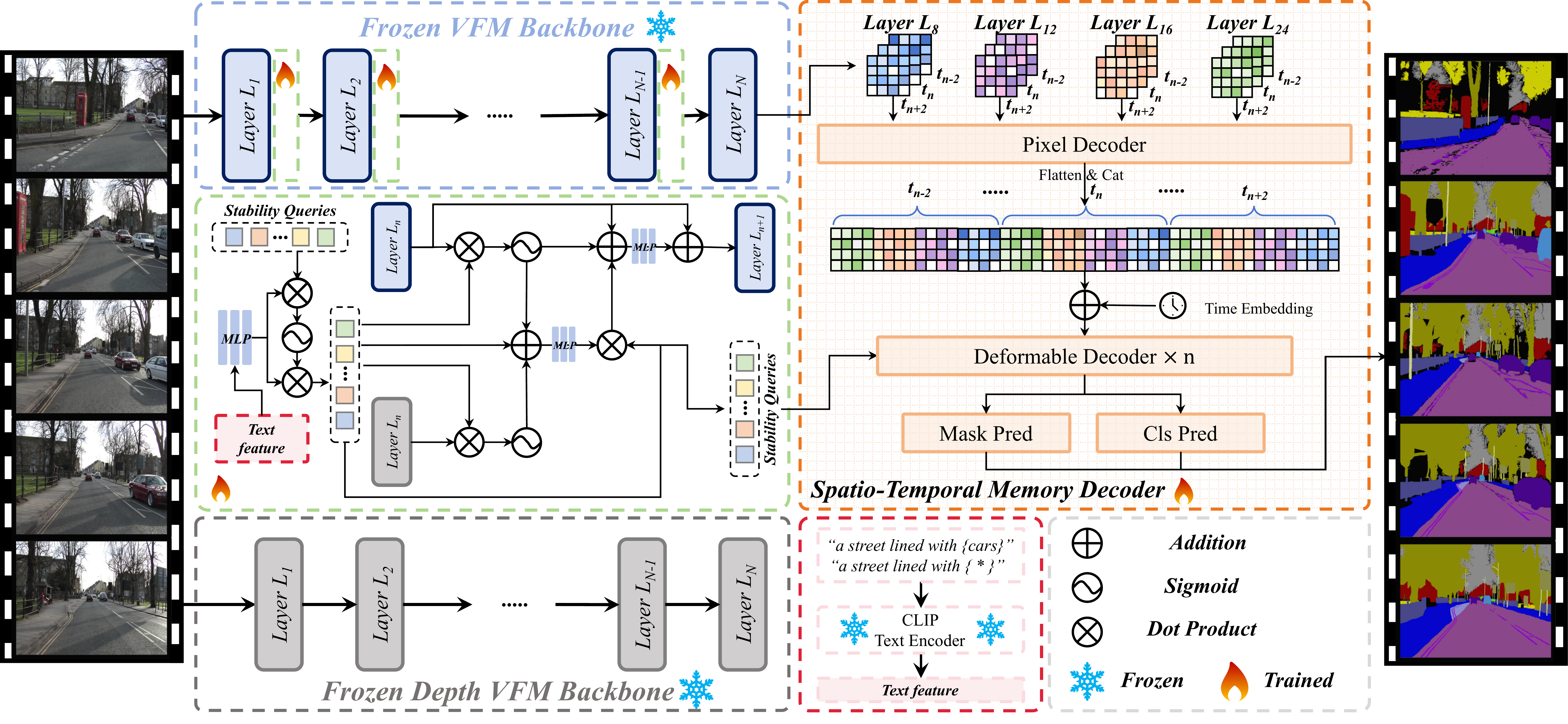}
  \setlength{\abovecaptionskip}{0.1cm}
  \setlength{\belowcaptionskip}{-0.2cm}
  \caption{Overview of the proposed \textbf{Time2General} framework with \textbf{Stability Queries} and \textbf{Spatio-Temporal Memory Decoder} for cross-frame consistent DGVSS.}
  \label{fig: Time2General}
\end{figure*}

\section{Related Work}

Video Semantic Segmentation (VSS) assigns pixel-level semantic labels to every frame and improves per-frame inference by exploiting temporal redundancy and motion cues \cite{ravi2024sam}. Propagation-based approaches warp and propagate semantics from reference frames using optical flow and learned motion fields with lightweight refinement \cite{zhu2017deep,jain2019accel,gadde2017semantic,karaev2024cotracker,liu2020efficient}, while recent variants further enhance consistency via semantic-similarity propagation and efficient temporal sharing mechanisms \cite{vincent2025high,hesham2025exploiting}. Clip-level methods aggregate multi-frame context within short clips through temporal attention, cross-frame affinity, and memory modules, and universal frameworks further promote video-level consistency across tasks \cite{wang2018non,miao2021vspw,wang2021temporal,cheng2022masked,shin2024video,zhang2025dvis++}.

Video Generalizable Semantic Segmentation (VGSS) targets robustness under distribution shift by jointly modeling temporal continuity and cross-domain generalization \cite{zhang2024video}. Prior unsupervised domain adaptation leverages unlabeled target domain data with temporal regularization \cite{shin2021unsupervised,xing2022domain}. Domain generalization methods avoid target data and are more practical when the target domain is unknown and dynamic \cite{bi2024learning}. Most VGSS methods extend DGSS strategies and introduce cross-frame consistency constraints (e.g., statistics and prototypes) to counter temporal drift under domain shifts \cite{wei2024stronger,lee2022wildnet,sakaridis2021acdc,huang2021cross,karaev2024cotracker}.

\section{Time2General}

\subsection{Overview}

Given a single labeled source video domain $\mathcal{D}_s$ for training, Domain-Generalized Video Semantic Segmentation (DGVSS) targets robust test performance on multiple unseen target video domains $\{\mathcal{D}_t^i\}$ without labels and without any test time adaptation. This setting poses three practical challenges: (1) \textbf{single-source and limited video supervision}, (2) \textbf{domain shift and degraded visibility}, and (3) \textbf{temporal sampling errors}.


To address these challenges, we propose \textbf{Time2General}, which aims to learn stability queries for temporally persistent semantic anchors. As shown in Figure~\ref{fig: Time2General}, we freeze a DINOv2 backbone to preserve the cross-domain prior and train only lightweight \textbf{Stability Queries} together with the \textbf{Spatio-Temporal Memory Decoder}. The Stability Queries (1) modulate and aggregate intermediate backbone features into robust multi-scale representations, mitigating overfitting under single-source supervision, and (2) provide a unified query space to integrate complementary generalization cues (e.g., geometric priors and text semantics).

Built upon query-conditioned pixel features, the Spatio-Temporal Memory Decoder models multi-frame context in a correspondence-free manner, avoiding explicit frame-to-frame matching. During training on $\mathcal{D}_s$, we further suppress flicker using the \textbf{Masked Temporal Consistency Loss} and randomized temporal-stride clip sampling to reduce sampling error.

\subsection{Stability Queries}

We introduce a set of learnable Stability Queries shared across all frames, serving as temporally persistent semantic anchors. To reduce overfitting under single-source supervision, we freeze the DINOv2 backbone and train only lightweight query-based modules with the segmentation decoder. Given the $t$-th frame, we extract intermediate token sequences from selected frozen DINOv2 blocks $l\in\mathcal{L}$: $\mathbf{X}^{(l)}_{t,\mathrm{rgb}}\in\mathbb{R}^{N^{(l)}\times C^{(l)}}$, where $N^{(l)}=H^{(l)}W^{(l)}$ is the number of spatial tokens and $C^{(l)}$ is the channel dimension.

We maintain $Q$ Stability Queries $\mathbf{S}\in\mathbb{R}^{Q\times d}$ in a shared query space of dimension $d$. For each layer $l$, we compute cross-attention $\mathbf{A}^{(l)}_{t,\mathrm{rgb}} =\mathrm{softmax}\!\left(\frac{\mathbf{Q}^{(l)}(\mathbf{K}^{(l)}_{t,\mathrm{rgb}})^\top}{\sqrt{d}}\right) \in\mathbb{R}^{Q\times N^{(l)}}$ from queries $\mathbf{Q}^{(l)}=\mathbf{S}\mathbf{W}_q^{(l)}$ to tokens $\mathbf{K}^{(l)}_{t,\mathrm{rgb}}=\mathbf{X}^{(l)}_{t,\mathrm{rgb}}\mathbf{W}_k^{(l)}, \mathbf{V}^{(l)}_{t,\mathrm{rgb}}=\mathbf{X}^{(l)}_{t,\mathrm{rgb}}\mathbf{W}_v^{(l)}$. Then, we obtain query-level summaries (e.g., semantic anchors) $\mathbf{U}^{(l)}_{t,\mathrm{qry,rgb}}
=\mathbf{A}^{(l)}_{t,\mathrm{rgb}}\mathbf{V}^{(l)}_{t,\mathrm{rgb}}\in\mathbb{R}^{Q\times d}$.

To modulate pixel features in the spirit of query-to-pixel interaction, we then project the updated queries back to spatial tokens using the same attention weights:
\begin{equation}
   \mathbf{Z}^{(l)}_{t,\mathrm{rgb}}
=\mathrm{Unflat}\!\left((\mathbf{A}^{(l)}_{t,\mathrm{rgb}})^\top \mathbf{U}^{(l)}_{t,\mathrm{qry,rgb}}\right),
\end{equation}
where $\mathrm{Unflat}(\cdot)$ reshapes a sequence of length $N^{(l)}$ into a spatial map of size $H^{(l)}\times W^{(l)}$. Finally, we combine the frozen backbone feature map with the query-induced modulation:
\begin{equation}
\mathbf{U}^{(l)}_{t,\mathrm{rgb}}
=\mathrm{LN}\!\left(\phi^{(l)}(\mathbf{F}^{(l)}_{t,\mathrm{rgb}}) + \mathbf{Z}^{(l)}_{t,\mathrm{rgb}}\right),
\end{equation}
where $\mathbf{F}^{(l)}_{t,\mathrm{rgb}}\in\mathbb{R}^{C^{(l)}\times H^{(l)}\times W^{(l)}}$ is the DINOv2 feature map, $\phi^{(l)}:\mathbb{R}^{C^{(l)}}\!\rightarrow\!\mathbb{R}^{d}$ denotes a lightweight $1\times1$ projection, and $\mathrm{LN}(\cdot)$ is LayerNorm.

We incorporate lightweight dataset-level priors from frozen text and depth encoders. Let $\mathbf{E}\in\mathbb{R}^{K\times d}$ denote the frozen class text embeddings and $\mathbf{e}\in\mathbb{R}^{d}$ a global semantic context vector; we bias the stability queries with an in-context as $\tilde{\mathbf{S}}=\mathbf{S}+\alpha_{\mathrm{txt}}\cdot\mathbf{e}$, where $\alpha_{\mathrm{txt}}$ is a learnable scalar, and we use $\tilde{\mathbf{S}}$ in place of $\mathbf{S}$ when computing $\mathbf{Q}^{(l)}$. In parallel, we extract frozen DepthAnything tokens $\mathbf{X}^{(l)}_{t,\mathrm{dep}}\in\mathbb{R}^{N^{(l)}\times C^{(l)}}$ for each selected block $l\in\mathcal{L}$; since the RGB and depth encoders share the same DINOv2 tokenization, $\mathbf{X}^{(l)}_{t,\mathrm{dep}}$ is spatially aligned with $\mathbf{X}^{(l)}_{t,\mathrm{rgb}}$ and requires no interpolation.

We compute an analogous query-to-token attention $\mathbf{Z}^{(l)}_{t,\mathrm{dep}}$ that is the same as the RGB branch. We fuse RGB and depth-induced modulations $\mathbf{Z}^{(l)}_{t}=\beta^{(l)}\,\mathbf{Z}^{(l)}_{t,\mathrm{rgb}} + (1-\beta^{(l)})\,\mathbf{Z}^{(l)}_{t,\mathrm{dep}}$ with a learnable layer-wise gate $\beta^{(l)}\in(0,1)$, and we obtain the final query-conditioned pixel feature map $\mathbf{U}^{(l)}_{t}=\mathrm{LN}\!\left(\phi^{(l)}(\mathbf{F}^{(l)}_{t,\mathrm{rgb}}) + \mathbf{Z}^{(l)}_{t}\right)$. The set $\{\mathbf{U}^{(l)}_{t}\}_{l\in\mathcal{L}}$ forms multi-scale, query-conditioned pixel features that are consumed by the subsequent spatio-temporal decoder.

\subsection{Spatio-Temporal Memory Decoder}

The Spatio-Temporal Memory Decoder aggregates multi-frame context in a correspondence-free manner by allowing the same Stability Queries to attend to a joint memory constructed from multi-frame, multi-scale query-conditioned pixel features. This design avoids explicit frame-to-frame matching and provides temporally consistent semantics under domain shift.

Given a clip of $T$ frames, for each frame $t$ and each selected scale $l$, we take the multi-scale query-conditioned pixel feature maps $\mathbf{U}^{(l)}_{t}\in\mathbb{R}^{B\times d\times H^{(l)}\times W^{(l)}}$ from the previous stage, where $B$ is the batch size and $d$ is the shared feature dimension. We flatten each scale map into a token sequence $\mathbf{M}^{(l)}_{t}=\mathrm{Flat}\!\left(\mathbf{U}^{(l)}_{t}\right)$ and add learnable temporal and scale embeddings $\hat{\mathbf{M}}^{(l)}_{t}=\mathbf{M}^{(l)}_{t}+\mathbf{e}_{\mathrm{tmp}}+\mathbf{e}_{\mathrm{scl}}^{(l)}$, where $\mathbf{e}_{\mathrm{tmp}}\in\mathbb{R}^{d}$ is indexed by the relative temporal position within the sampled clip, and $\mathbf{e}_{\mathrm{scl}}^{(l)}\in\mathbb{R}^{d}$ encodes the scale identity. Both are broadcast to all $N^{(l)}$ spatial tokens.

We concatenate tokens across time and scales to form the joint memory, and then we broadcast Stability Queries to the batch dimension and update them by attending to $\mathbf{M}$ through $N_{\mathrm{dec}}$ stacked decoder blocks:
\begin{equation}
    \mathbf{M}=\bigoplus_{t=1}^{T}\ \bigoplus_{l\in\mathcal{L}}\ \hat{\mathbf{M}}^{(l)}_{t}
\in\mathbb{R}^{B\times N\times d}, N=\sum_{t=1}^{T}\sum_{l\in\mathcal{L}} N^{(l)},
\end{equation}
\begin{equation}
\mathbf{S}^{(k+1)} = \mathrm{FFN}\!\left(
\mathbf{S}^{(k)}+\mathrm{Defor. Attn.}\!\left(\mathbf{S}^{(k)},\mathbf{M}\right)\right),
\end{equation}
where $k=0,\ldots,N_{\mathrm{dec}}-1$, $\mathbf{S}^{(0)}=\mathbf{S}$. The final temporally-aware queries are
$\bar{\mathbf{S}}=\mathbf{S}^{(N_{\mathrm{dec}})}\in\mathbb{R}^{Q\times d}$.

For each frame $t$, we fuse its multi-scale query-conditioned pixel features into a single per-pixel representation:
\begin{equation} 
\mathbf{P}_t=\mathrm{Fuse}\!\left(\left\{\mathbf{U}^{(l)}_{t}\right\}_{l\in\mathcal{L}}\right)\in\mathbb{R}^{B\times d\times H\times W},
\end{equation}
where $\mathrm{Fuse}(\cdot)$ is a lightweight, multi-scale fusion module. Conditioned on $\bar{\mathbf{S}}$, we predict segmentation logits for frame $t$ with a query-driven head:
\begin{equation}
    \hat{\mathbf{Y}}_{t}=g\!\left(\mathbf{P}_t,\bar{\mathbf{S}}\right)\in\mathbb{R}^{B\times K\times H\times W},
\end{equation}
where $K$ is the number of semantic classes, and $g(\cdot)$ denotes the segmentation prediction module.

We apply random-stride clip sampling to improve robustness to temporal sampling shifts: for each training clip, we pick a start index $u$ and sample a stride $r\in\mathcal{R}$ once, then form the clip by uniform sub-sampling ${I_{u+(t-1)r}}{t=1}^{T}$ so that all adjacent gaps are identical; temporal embeddings $\mathbf{e}_{\mathrm{tmp}}^{(t)}$ are indexed by the relative position $t$ within the sampled clip, and varying $r$ across clips exposes the model to diverse temporal gaps during training, thereby improving robustness to test-time sampling-rate changes.

\begin{table*}[!t]
  \centering
  \caption{Performance comparison between our Time2General and existing DGSS and VSS methods under \textit{CamVid, Apollo., KITTI-360 $\rightarrow$ City.-s-C}. (\%) }
  \label{tab:cityscapes}
  \resizebox{\textwidth}{!}{%
  \begin{tabular}{c c *{3}{c} *{3}{c} *{3}{c} *{3}{c}}
    \toprule
    \multirow{2}{*}{Method}
    & \multirow{2}{*}{Proc.\ \& Year}
    & \multicolumn{3}{c}{Cityscapes-Fog}
    & \multicolumn{3}{c}{Cityscapes-Frost}
    & \multicolumn{3}{c}{Cityscapes-Snow}
    & \multicolumn{3}{c}{Cityscapes-Spatter} \\
    \cmidrule(lr){3-5}\cmidrule(lr){6-8}\cmidrule(lr){9-11}\cmidrule(lr){12-14}
    & & mIoU & mVC$_8$ & mVC$_{16}$
      & mIoU & mVC$_8$ & mVC$_{16}$
      & mIoU & mVC$_8$ & mVC$_{16}$
      & mIoU & mVC$_8$ & mVC$_{16}$ \\
      \midrule

      \multicolumn{14}{c}{\textbf{Train on KITTI-360}} \\
    \multicolumn{2}{l}{\textit{DGSS Method:}} \\
    REIN~\cite{wei2024stronger}
      & CVPR2024
      & 40.54 & 58.82 & 32.66
      & 28.32 & 37.92 & 27.92
      & 38.29 & 79.14 & 77.68
      & 41.19 & 80.59 & 79.62 \\
    FADA~\cite{bi2024learning}
      & NeurIPS2024
      & 45.84 & 53.08 & 30.05
      & 30.00 & 37.65 & 33.12
      & 41.00 & 74.04 & 72.62
      & 43.04 & 73.35 & 69.99 \\
    DepthForge~\cite{chen2025stronger}
      & ICCV2025
      & 46.77 & 74.95 & 73.45
      & 30.41 & 36.42 & 28.00
      & 36.50 & 74.95 & 73.45
      & 36.45 & 73.58 & 71.40 \\
    \multicolumn{2}{l}{\textit{VSS Method:}} \\
    SSP~\cite{vincent2025high}
      & CVPR2025
      & 25.03 & 53.22 & 43.13
      & 10.10 & 19.93 & 15.65
      & 11.97 & 32.63 & 32.89
      & 16.60 & 35.13 & 32.31  \\
    TV3S~\cite{hesham2025exploiting}
      & CVPR2025
      & 6.85 & 10.89 & 7.18
      & 4.24 & 8.33 & 6.82
      & 8.77 & 19.40 & 18.85
      & 3.78 & 19.71 & 20.05 \\
    \multicolumn{2}{l}{\textit{DGVSS Method:}} \\
    \textbf{Time2General (Ours)}
      & -
      & 49.27 & 80.43 & 79.78
      & 32.25 & 41.25 & 33.45
      & 44.62 & 83.31 & 83.89
      & 48.62 & 85.10 & 85.50 \\
    \textit{Improv.\ vs.\ Best}
      &  & $+2.50\uparrow$ & $+5.48\uparrow$ & $+6.33\uparrow$
      & $+2.25\uparrow$ & $+3.33\uparrow$ & $+0.33\uparrow$
      & $+3.62\uparrow$ & $+4.17\uparrow$ & $+6.21\uparrow$
      & $+5.58\uparrow$ & $+4.51\uparrow$ & $+5.88\uparrow$ \\
    \textit{Improv.\ vs.\ Best VSS}
      &  & \textbf{+24.24 $\uparrow$} & \textbf{+27.21 $\uparrow$} & \textbf{+36.65 $\uparrow$}
      & \textbf{+22.15 $\uparrow$} & \textbf{+21.32 $\uparrow$} & \textbf{+17.80 $\uparrow$} & \textbf{+28.97 $\uparrow$} & \textbf{+50.68 $\uparrow$} & \textbf{+51.00 $\uparrow$}
      & \textbf{+32.02 $\uparrow$} & \textbf{+49.97 $\uparrow$} & \textbf{+53.19 $\uparrow$} \\
      \midrule

      \multicolumn{14}{c}{\textbf{Train on ApolloScape}} \\
    \multicolumn{2}{l}{\textit{DGSS Method:}} \\
    REIN~\cite{wei2024stronger}
      & CVPR2024
      & 38.92 & 59.08 & 32.74
      & 27.64 & 33.45 & 23.34
      & 35.94 & 65.37 & 65.10
      & 35.22 & 68.89 & 68.10 \\
    FADA~\cite{bi2024learning}
      & NeurIPS2024
      & 35.84 & 53.08 & 28.05
      & 30.00 & 37.65 & 23.12
      & 34.00 & 64.04 & 62.62
      & 36.04 & 63.35 & 62.99 \\
    DepthForge~\cite{chen2025stronger}
      & ICCV2025
      & 41.75 & 62.02 & 35.56
      & 31.10 & 36.56 & 28.56
      & 28.96 & 60.77 & 51.74
      & 40.52 & 63.83 & 59.39 \\
    \multicolumn{2}{l}{\textit{VSS Method:}} \\
    SSP~\cite{vincent2025high}
      & CVPR2025
      & 26.73 & 63.24 & 58.32
      & 15.93 & 33.85 & 27.04
      & 22.04 & 68.21 & 62.67
      & 15.46 & 68.30 & 68.21 \\
    TV3S~\cite{hesham2025exploiting}
      & CVPR2025
      & 11.38 & 43.94 & 42.33
      & 10.25 & 20.05 & 15.06
      & 11.43 & 41.84 & 33.69
      & 12.03 & 67.12 & 68.52 \\
    \multicolumn{2}{l}{\textit{DGVSS Method:}} \\
    \textbf{Time2General (Ours)}
      & -
      & 43.30 & 73.46 & 70.85
      & 31.50 & 41.14 & 32.72
      & 37.23 & 71.79 & 71.08
      & 41.58 & 71.81 & 70.23 \\
    \textit{Improv.\ vs.\ Best}
      &  & $+1.55\uparrow$ & $+10.22\uparrow$ & $+12.53\uparrow$
      & $+0.40\uparrow$ & $+3.49\uparrow$ & $+4.16\uparrow$
      & $+1.29\uparrow$ & $+3.58\uparrow$ & $+5.98\uparrow$
      & $+1.06\uparrow$ & $+2.92\uparrow$ & $+1.71\uparrow$ \\
    \textit{Improv.\ vs.\ Best VSS}
      &  & \textbf{+16.57 $\uparrow$} & \textbf{+10.22 $\uparrow$} & \textbf{+12.53 $\uparrow$}
      & \textbf{+15.57 $\uparrow$} & \textbf{+7.29 $\uparrow$} & \textbf{+5.68 $\uparrow$}
      & \textbf{+10.19 $\uparrow$} & \textbf{+3.58 $\uparrow$} & \textbf{+8.41 $\uparrow$}
      & \textbf{+26.12 $\uparrow$} & \textbf{+3.51 $\uparrow$} & \textbf{+1.71 $\uparrow$} \\
      \midrule

      \multicolumn{14}{c}{\textbf{Train on CamVid}} \\
    \multicolumn{2}{l}{\textit{DGSS Method:}} \\
    REIN~\cite{wei2024stronger}
      & CVPR2024
      & 42.53 & 50.92 & 49.69
      & 29.87 & 30.42 & 27.60
      & 33.41 & 68.85 & 63.84
      & 39.24 & 65.21 & 37.09 \\
    FADA~\cite{bi2024learning}
      & NeurIPS2024
      & 42.65 & 38.88 & 18.79
      & 28.62 & 32.20 & 22.29
      & 24.75 & 44.11 & 45.35
      & 35.68 & 55.83 & 50.93 \\
    DepthForge~\cite{chen2025stronger}
      & ICCV2025
      & 43.61 & 54.20 & 36.34
      & 28.98 & 33.51 & 27.53
      & 36.99 & 66.91 & 64.55
      & 40.90 & 72.88 & 66.96 \\
    \multicolumn{2}{l}{\textit{VSS Method:}} \\
    SSP~\cite{vincent2025high}
      & CVPR2025
      & 39.57 & 50.79 & 44.37
      & 21.33 & 35.80 & 21.83
      & 25.19 & 55.44 & 56.70
      & 36.50 & 61.51 & 60.68 \\
    TV3S~\cite{hesham2025exploiting}
      & CVPR2025
      & 16.21 & 26.38 & 17.44
      & 15.93 & 31.08 & 25.79
      & 19.09 & 47.81 & 48.97
      & 19.36 & 47.48 & 44.65 \\
    \multicolumn{2}{l}{\textit{DGVSS Method:}} \\
    \textbf{Time2General (Ours)}
      & -
      & 45.61 & 61.87 & 53.19
      & 31.28 & 36.90 & 29.59
      & 39.29 & 70.47 & 66.04
      & 45.74 & 80.91 & 78.63 \\
    \textit{Improv.\ vs.\ Best}
      &  & $+2.00\uparrow$ & $+7.67\uparrow$ & $+3.50\uparrow$
      & $+1.41\uparrow$ & $+1.10\uparrow$ & $+1.99\uparrow$
      & $+2.30\uparrow$ & $+1.62\uparrow$ & $+1.99\uparrow$
      & $+4.84\uparrow$ & $+8.03\uparrow$ & $+11.67\uparrow$ \\
    \textit{Improv.\ vs.\ Best VSS}
      &  & \textbf{+6.04 $\uparrow$} & \textbf{+11.08 $\uparrow$} & \textbf{+8.82 $\uparrow$}
      & \textbf{+9.95 $\uparrow$} & \textbf{+1.10 $\uparrow$} & \textbf{+3.80 $\uparrow$}
      & \textbf{+14.10 $\uparrow$} & \textbf{+15.03 $\uparrow$} & \textbf{+9.34 $\uparrow$}
      & \textbf{+9.24 $\uparrow$} & \textbf{+16.40 $\uparrow$} & \textbf{+17.95 $\uparrow$} \\
    \bottomrule
  \end{tabular}
  }
\end{table*}

\subsection{Masked Temporal Consistency Loss}

Following the domain generalization protocol, our temporal regularizer is computed only on source-domain training clips with ground-truth labels. We propose the Masked Temporal Consistency Loss to suppress frame-to-frame flicker by penalizing prediction changes only in stable regions. The loss further improves robustness to temporal-sampling shifts via multi-stride temporal differences and robust trimming.

Let $\mathbf{x}\in\mathbb{R}^{B\times T\times K\times H\times W}$ be the per-frame logits and
$\mathbf{y}\in\{0,\dots,K-1\}$ be the ground-truth labels. We convert logits to probabilities $\mathbf{p}=\mathrm{softmax}(\mathbf{x})$ along the class dimension. For the temporal scale $s\in\{0,\dots,S-1\}$, we use a stride of $r_s=2^s$ and compare frames separated by $r_s$:
\begin{equation}
    \delta^{(s)}_{b,t}(u)
=\left\lVert \mathbf{p}_{b,t+r_s}(:,u)-\mathbf{p}_{b,t}(:,u)\right\rVert_1,
\end{equation}
where $u$ indexes spatial locations and $\lVert\cdot\rVert_1$ sums absolute differences over the $K$ classes.

We apply the penalty only when the ground-truth label remains unchanged and is valid:
\begin{equation}
    \begin{aligned}
v^{(s)}_{b,t}(u) 
&= \mathbb{I}\!\left[\mathbf{y}_{b,t}(u)\neq \iota\right]
   \cdot \mathbb{I}\!\left[\mathbf{y}_{b,t+r_s}(u)\neq \iota\right],\\
m^{(s)}_{b,t}(u)
&= v^{(s)}_{b,t}(u)\cdot
   \mathbb{I}\!\left[\mathbf{y}_{b,t}(u)=\mathbf{y}_{b,t+r_s}(u)\right],
\end{aligned}
\end{equation}
We collect temporal differences over stable pixels:
\begin{equation}
    \mathcal{V}^{(s)}=\left\{\delta^{(s)}_{b,t}(u)\ \middle|\ m^{(s)}_{b,t}(u)=1\right\},
\end{equation}

To reduce the influence of noisy pixels (e.g., boundaries and uncertain regions), we compute a trimmed mean over $\mathcal{V}^{(s)}$. Given the trimming ratio $\tau\in[0,1)$, we keep the smallest $(1-\tau)$ fraction of values in $\mathcal{V}^{(s)}$ and define:
\begin{equation}
    \mathrm{TrimMean}_\tau(\mathcal{V}^{(s)})=\frac{1}{|\hat{\mathcal{V}}^{(s)}|}\sum_{v\in\hat{\mathcal{V}}^{(s)}} v,
\end{equation}
where $\hat{\mathcal{V}}^{(s)}\subseteq\mathcal{V}^{(s)}$ contains the lowest $(1-\tau)|\mathcal{V}^{(s)}|$ values. We weight each temporal scale with a decay factor $\alpha^s$ and average over the valid scales:
\begin{equation}
\mathcal{L}_{\mathrm{mtc}}=\frac{1}{|\mathcal{S}|}\sum_{s\in\mathcal{S}}\alpha^s\cdot \mathrm{TrimMean}_\tau(\mathcal{V}^{(s)}).
\end{equation}
where $\mathcal{S}$ is the set of valid scales with a non-empty $\mathcal{V}^{(s)}$. The final loss is $\lambda_{\mathrm{mtc}}\mathcal{L}_{\mathrm{mtc}}$.

For the proposed MTC loss, we set the trimming ratio to $\tau=0.2$, and adopt a decay factor $\alpha=0.5$ in Eq.~11. The loss weight $\lambda_{\mathrm{mtc}}$ is set to 1 by default.

\section{Experiment}

\begin{figure*}[!t]
  \centering
  \includegraphics[width=0.95\textwidth]{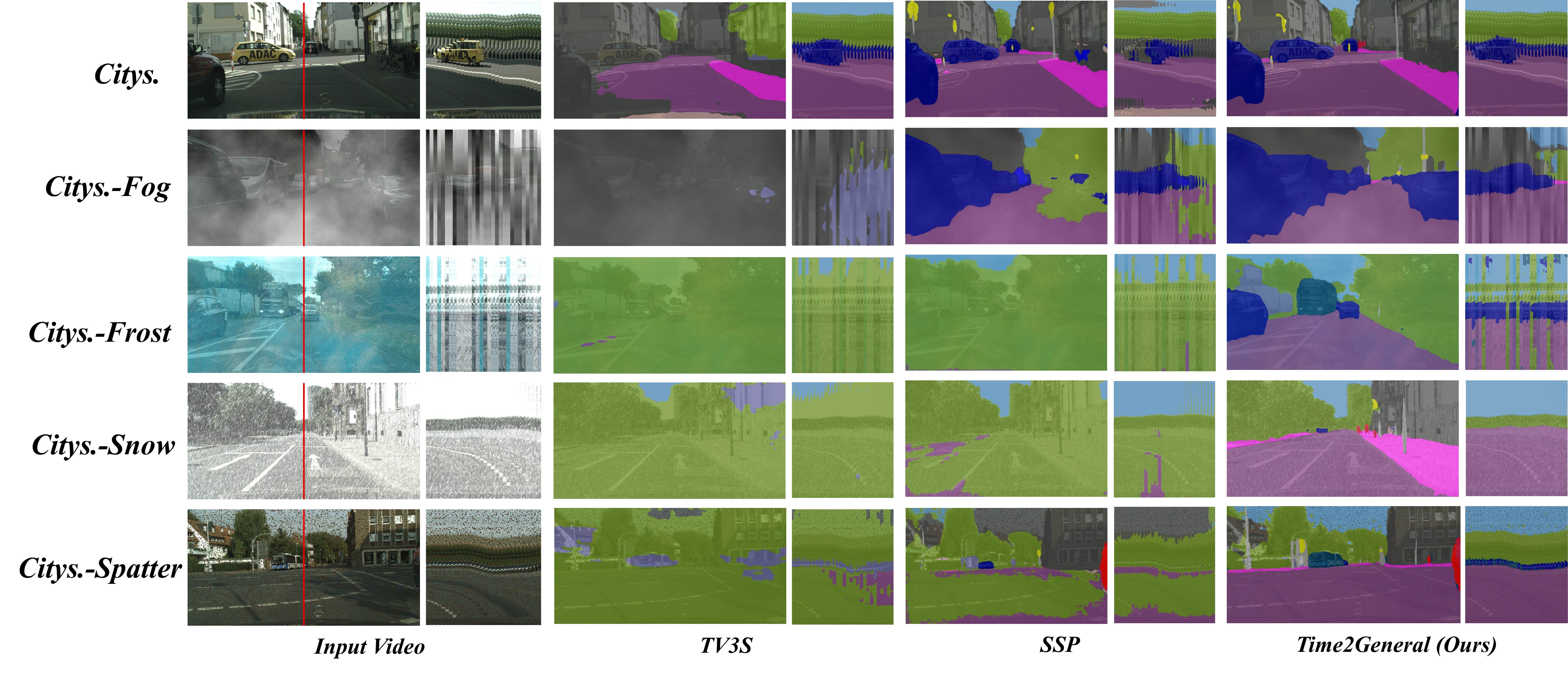}
  \setlength{\abovecaptionskip}{-0.1cm}
  \setlength{\belowcaptionskip}{-0.2cm}
  \caption{Qualitative comparison of cross domain segmentation predictions for \textit{KITTI-360 $\rightarrow$ City.-s + City.-s-C}.}
  \label{fig: kitti2city}
\end{figure*}

\subsection{Datasets \& Evaluation Protocols}

\textbf{Datasets and Label Alignment.} We evaluate our method on five well-known driving video datasets, including \textbf{KITTI-360}, \textbf{ApolloScape}, \textbf{CamVid}, \textbf{Cityscapes-sequence}, and \textbf{Cityscapes-sequence-Corrupted}. We further provide label alignment across all datasets and evaluate on \textbf{15 classes}. The detailed dataset descriptions and label settings are provided in the appendix~\ref{app: label}.

In our benchmark, different datasets exhibit noticeably different effective temporal sampling densities. In particular, KITTI-360 and Apollo. provide long driving sequences with denser temporal supervision, while CamVid and Citys.-s consist of shorter clips/snippets with sparser temporal coverage. We treat cross-dataset transfers between these groups as a practical instance of temporal-sampling shift.

\textbf{Metrics.} We evaluate segmentation accuracy using mIoU and temporal stability using mean Video Consistency ($\mathrm{mVC}_n$) with a window size $n$, reported under (1) dense-GT and (2) sparse-GT evaluation using static-pixel masks. The detailed definitions are provided in the appendix~\ref{app: metrics}. For Cityscapes sequence and its corrupted variant, mVC is evaluated with sparse GT (approx.) protocol; for other datasets, mVC is evaluated with dense GT protocol.

\textbf{Evaluation protocol.} Following the standard domain generalization protocol, we train on a single source dataset and evaluate on multiple unseen target domains. We consider three setups:
\begin{itemize}
    \item KITTI-360 $\rightarrow$ CamVid, Apollo., Citys.-s, Citys.-s-C;
    \item Apollo. $\rightarrow$ CamVid, KITTI-360, Citys.-s, Citys.-s-C;
    \item CamVid $\rightarrow$ Apollo., KITTI-360, Citys.-s, Citys.-s-C.
\end{itemize}

\subsection{Implementation Details}

Our implementation is built on MMSegmentation \cite{mmseg2020} and optimized with AdamW (initial learning rate $6\times10^{-5}$, weight decay $0.05$, $\epsilon{=}10^{-8}$, $(\beta_1,\beta_2){=}(0.9,0.999)$) under a polynomial learning rate schedule that decays to zero without warmup. We train for 30{,}000 iterations for all source domains (KITTI-360, Apollo., and CamVid). For densely sampled sources (KITTI-360), we additionally randomize the training stride by sampling a single clip-level interval $r\in\{5,10,15,20,30,40\}$, which is fixed within the clip and used only during training. The depth representation is extracted by the same frozen Depth Anything backbone as in DepthForge, introducing no additional supervision or training overhead. We use \emph{Stability Queries} whose number and query dimension match the learnable tokens in DepthForge. Data augmentation includes multi-scale resizing, random cropping with a fixed crop size and category-ratio constraint, random horizontal flipping, and photometric distortion. All experiments are run on a single NVIDIA RTX PRO 6000 Blackwell Server Edition with batch size 4. For inference, to ensure fair comparison and efficiency, our current implementation concatenates clip predictions without overlap, and we disable temporal stride randomization at test time.

\begin{table*}[ht]
  \centering
  \caption{Performance comparison between our Time2General and existing DGSS and VSS methods under \textit{KITTI-360 $\rightarrow$ CamVid + Apollo. + City.-s + City.-s-C} generalization settings. (\%)}
  \label{tab:KITTI-360}
  \resizebox{\textwidth}{!}{%
  \begin{tabular}{c c *{3}{c} *{3}{c} *{3}{c} *{3}{c}}
    \toprule
    \multirow{2}{*}{Method}
    & \multirow{2}{*}{Proc.\ \& Year}
    & \multicolumn{3}{c}{CamVid}
    & \multicolumn{3}{c}{Apollo.}
    & \multicolumn{3}{c}{Citys.-s}
    & \multicolumn{3}{c}{Citys.-s-C} \\
    \cmidrule(lr){3-5}\cmidrule(lr){6-8}\cmidrule(lr){9-11}\cmidrule(lr){12-14}
    & & mIoU & mVC$_8$ & mVC$_{16}$
      & mIoU & mVC$_8$ & mVC$_{16}$
      & mIoU & mVC$_8$ & mVC$_{16}$
      & mIoU & mVC$_8$ & mVC$_{16}$ \\
      \midrule
      \multicolumn{2}{l}{\textit{DGSS Method:}} \\
    REIN~\cite{wei2024stronger} 
      & CVPR2024
      & 41.19
      & 84.03
      & 83.53
      & 38.92
      & 81.84
      & 83.36
      & 53.69
      & 86.96
      & 87.27
      & 37.09
      & 64.12
      & 54.47 \\
    FADA~\cite{bi2024learning} 
      & NeurIPS2024
      & 55.51
      & 83.07
      & 82.00
      & 44.42
      & 81.42
      & 82.57
      & 47.76
      & 74.71
      & 74.29
      & 39.97
      & 59.53
      & 51.45 \\
    DepthForge~\cite{chen2025stronger}
      & ICCV2025
      & 51.28
      & 84.65
      & 84.26
      & 43.87
      & 82.90
      & 84.11
      & 54.72
      & 87.00
      & 87.34
      & 37.53
      & 64.98
      & 61.58 \\
    
    \multicolumn{2}{l}{\textit{VSS Method:}} \\
    SSP~\cite{vincent2025high} 
      & CVPR2025 
      &  48.42
      &  90.39
      &  91.43
      &  31.90
      &  84.86
      &  85.22
      &  53.20
      &  87.56
      &  87.25
      &  15.93
      &  35.23
      &  31.00\\
    TV3S~\cite{hesham2025exploiting}
      & CVPR2025
      &  48.74
      &  86.18
      &  85.61
      &  22.43
      &  75.80
      &  75.16
      &  23.23
      &  59.22
      &  57.12
      &  5.91
      &  14.58
      &  13.23\\
      \multicolumn{2}{l}{\textit{DGVSS Method:}} \\
    \textbf{Time2General (Ours)} 
      & -
      & 72.09
      & 94.56
      & 94.76
      & 50.51
      & 91.73
      & 94.76
      & 55.18
      & 88.87
      & 89.07
      & 43.69
      & 72.52
      & 70.66\\
    \textit{Improv.\ vs.\ Best} &  & $+16.58\uparrow$ & $+4.17\uparrow$ & $+3.33\uparrow$
        & $+6.09\uparrow$ & $+6.87\uparrow$ & $+9.54\uparrow$
        & $+0.46\uparrow$ & $+1.31\uparrow$ & $+1.73\uparrow$
        & $+3.72\uparrow$ & $+7.54\uparrow$ & $+9.08\uparrow$ \\
    \textit{Improv.\ vs.\ Best VSS} &  & \textbf{+23.35 $\uparrow$} & \textbf{+4.17 $\uparrow$} & \textbf{+3.33 $\uparrow$}  & \textbf{+18.61 $\uparrow$} & \textbf{+6.87 $\uparrow$} & \textbf{+9.54 $\uparrow$}
    & \textbf{+1.98 $\uparrow$} & \textbf{+1.31 $\uparrow$} & \textbf{+1.82 $\uparrow$} & \textbf{+27.76 $\uparrow$} & \textbf{+37.29 $\uparrow$} & \textbf{+39.66 $\uparrow$} \\
    \bottomrule
  \end{tabular}
  }
\end{table*}

\begin{table*}[!t]
  \centering
  \caption{Performance comparison between our Time2General and existing DGSS and VSS methods under \textit{Apollo. $\rightarrow$ CamVid + KITTI-360 + City.-s + City.-s-C} generalization settings. (\%)}
  \label{tab:apollo}
  \resizebox{\textwidth}{!}{%
  \begin{tabular}{c c *{3}{c} *{3}{c} *{3}{c} *{3}{c}}
    \toprule
    \multirow{2}{*}{Method}
    & \multirow{2}{*}{Proc.\ \& Year}
    & \multicolumn{3}{c}{CamVid}
    & \multicolumn{3}{c}{KITTI-360}
    & \multicolumn{3}{c}{Citys.-s}
    & \multicolumn{3}{c}{Citys.-s-C} \\
    \cmidrule(lr){3-5}\cmidrule(lr){6-8}\cmidrule(lr){9-11}\cmidrule(lr){12-14}
    & & mIoU & mVC$_8$ & mVC$_{16}$
      & mIoU & mVC$_8$ & mVC$_{16}$
      & mIoU & mVC$_8$ & mVC$_{16}$
      & mIoU & mVC$_8$ & mVC$_{16}$ \\
      \midrule
      \multicolumn{2}{l}{\textit{DGSS Method:}} \\
    REIN~\cite{wei2024stronger} 
      & CVPR2024
      & 44.38
      & 67.69
      & 68.03
      & 35.83
      & 69.94
      & 67.96
      & 33.02
      & 75.25
      & 74.36
      & 34.43
      & 56.70
      & 47.32  \\
    FADA~\cite{bi2024learning} 
      & NeurIPS2024
      & 19.23
      & 63.07
      & 62.00
      & 39.79
      & 62.98
      & 62.12
      & 37.76
      & 74.71
      & 74.29
      & 33.97
      & 54.53
      & 44.20 \\
    DepthForge~\cite{chen2025stronger}
      & ICCV2025
      & 48.61
      & 71.60
      & 72.25
      & 45.05
      & 75.25
      & 74.36
      & 39.30
      & 76.46
      & 76.25
      & 35.58
      & 55.80
      & 43.81 \\
    \multicolumn{2}{l}{\textit{VSS Method:}} \\
    SSP~\cite{vincent2025high} 
      & CVPR2025 
      & 32.85
      & 67.40
      & 69.05
      & 26.42
      & 77.26
      & 68.47
      & 39.98
      & 71.92
      & 72.56
      & 20.04
      & 58.40
      & 54.06 \\
    TV3S~\cite{hesham2025exploiting}
      & CVPR2025
      & 23.05
      & 54.26
      & 58.67
      & 16.47
      & 64.46
      & 63.59
      & 20.38
      & 73.78
      & 74.98
      & 11.27
      & 43.24
      & 39.90  \\
      \multicolumn{2}{l}{\textit{DGVSS Method:}} \\
    \textbf{Time2General (Ours)} 
      & -
      & 59.70
      & 73.80
      & 74.03
      & 47.71
      & 77.51
      & 75.94
      & 47.08
      & 79.16
      & 78.00
      & 38.40
      & 64.55
      & 61.22\\
    \textit{Improv.\ vs.\ Best} &  & $+11.09\uparrow$ & $+2.20\uparrow$ & $+1.78\uparrow$
        & $+2.66\uparrow$ & $+0.25\uparrow$ & $+1.58\uparrow$
        & $+7.10\uparrow$ & $+2.70\uparrow$ & $+1.75\uparrow$
        & $+2.82\uparrow$ & $+6.15\uparrow$ & $+7.16\uparrow$ \\
    \textit{Improv.\ vs.\ Best VSS} &  & \textbf{+26.85 $\uparrow$} & \textbf{+6.40 $\uparrow$} & \textbf{+4.98 $\uparrow$}  & \textbf{+21.29 $\uparrow$} & \textbf{+0.25 $\uparrow$} & \textbf{+7.47 $\uparrow$} & \textbf{+7.10 $\uparrow$} & \textbf{+5.38 $\uparrow$} & \textbf{+3.02 $\uparrow$} & \textbf{+18.36 $\uparrow$} & \textbf{+6.15 $\uparrow$} & \textbf{+7.16 $\uparrow$} \\
    \bottomrule
  \end{tabular}
  }
\end{table*}

\begin{table*}[!t]
  \centering
  \caption{Performance comparison between our Time2General and existing DGSS and VSS methods under \textit{CamVid $\rightarrow$ Apollo. + KITTI-360 + City.-s + City.-s-C} generalization settings. (\%)}
  \label{tab:CamVid}
  \resizebox{\textwidth}{!}{%
  \begin{tabular}{c c *{3}{c} *{3}{c} *{3}{c} *{3}{c}}
    \toprule
    \multirow{2}{*}{Method}
    & \multirow{2}{*}{Proc.\ \& Year}
    & \multicolumn{3}{c}{Apollo.}
    & \multicolumn{3}{c}{KITTI-360}
    & \multicolumn{3}{c}{Citys.-s}
    & \multicolumn{3}{c}{Citys.-s-C} \\
    \cmidrule(lr){3-5}\cmidrule(lr){6-8}\cmidrule(lr){9-11}\cmidrule(lr){12-14}
    & & mIoU & mVC$_8$ & mVC$_{16}$
      & mIoU & mVC$_8$ & mVC$_{16}$
      & mIoU & mVC$_8$ & mVC$_{16}$
      & mIoU & mVC$_8$ & mVC$_{16}$ \\
      \midrule
      \multicolumn{2}{l}{\textit{DGSS Method:}} \\
    REIN~\cite{wei2024stronger} 
      & CVPR2024
      & 41.21
      & 71.60
      & 72.25
      & 35.83
      & 75.25
      & 74.36
      & 49.30
      & 78.15
      & 76.75
      & 36.26
      & 53.85
      & 44.56 \\
    FADA~\cite{bi2024learning} 
      & NeurIPS2024
      & 39.94
      & 88.46
      & 89.11
      & 48.15
      & 84.89
      & 80.56
      & 48.99
      & 84.50
      & 83.96
      & 32.93
      & 42.76
      & 34.34 \\
    DepthForge~\cite{chen2025stronger}
      & ICCV2025
      & 44.38
      & 80.38
      & 81.59
      & 54.97
      & 89.12
      & 87.04
      & 53.72
      & 87.23
      & 87.21
      & 37.62
      & 56.88
      & 48.85 \\
      \midrule
    \multicolumn{2}{l}{\textit{VSS Method:}} \\
    SSP~\cite{vincent2025high} 
      & CVPR2025 
      & 39.38
      & 82.09
      & 83.75
      & 52.60
      & 89.54
      & 88.56
      & 40.67
      & 87.92
      & 87.54
      & 30.65
      & 50.89
      & 45.90 \\ 
    TV3S~\cite{hesham2025exploiting}
      & CVPR2025
      & 22.64
      & 77.20
      & 77.42
      & 28.22
      & 70.50
      & 64.80
      & 36.52
      & 81.79
      & 81.92
      & 17.65
      & 38.19
      & 34.21 \\ 
      \midrule
      \multicolumn{2}{l}{\textit{DGVSS Method:}} \\
    \textbf{Time2General (Ours)} 
      & -
      & 47.66
      & 90.68
      & 90.92
      & 58.40
      & 90.44
      & 89.13
      & 55.11
      & 89.82
      & 88.58
      & 40.48
      & 62.53
      & 56.86\\
    \textit{Improv.\ vs.\ Best} &  & $+3.28\uparrow$ & $+2.22\uparrow$ & $+1.81\uparrow$
        & $+3.43\uparrow$ & $+0.90\uparrow$ & $+0.57\uparrow$
        & $+1.39\uparrow$ & $+1.90\uparrow$ & $+1.04\uparrow$
        & $+2.86\uparrow$ & $+5.65\uparrow$ & $+8.01\uparrow$ \\
    \textit{Improv.\ vs.\ Best VSS} &  & \textbf{+8.28 $\uparrow$} & \textbf{+8.59 $\uparrow$} & \textbf{+7.17 $\uparrow$} & \textbf{+5.80 $\uparrow$} & \textbf{+0.90 $\uparrow$} & \textbf{+0.57 $\uparrow$} & \textbf{+14.44 $\uparrow$} & \textbf{+1.90 $\uparrow$} & \textbf{+1.04 $\uparrow$}  & \textbf{+9.83 $\uparrow$} & \textbf{+11.64 $\uparrow$} & \textbf{+10.96 $\uparrow$} \\
    \bottomrule
  \end{tabular}
  }
\end{table*}

\subsection{Performance Comparison}

\textbf{Generalization to adverse weather corruptions.} Table~\ref{tab:cityscapes} reports results for models trained on KITTI-360, ApolloScape, and CamVid, evaluated on Cityscapes with four adverse weather corruptions. Time2General achieves the best performance across all corruptions and training sources. The larger margin over video baselines indicates stronger robustness to visibility degradation and temporal perturbations. Figure~\ref{fig: kitti2city} provides more qualitative evidence. Across different conditions, SSP and TV3S often suffer from texture-driven false positives that fluctuate across frames. In contrast, Time2General preserves more coherent and plausible object extents while maintaining stable labels on moving objects. These visual trends are consistent with our motivation, indicating stronger temporal coherence under reduced visibility.

\begin{figure*}[!t]
  \centering
  \includegraphics[width=0.95\textwidth]{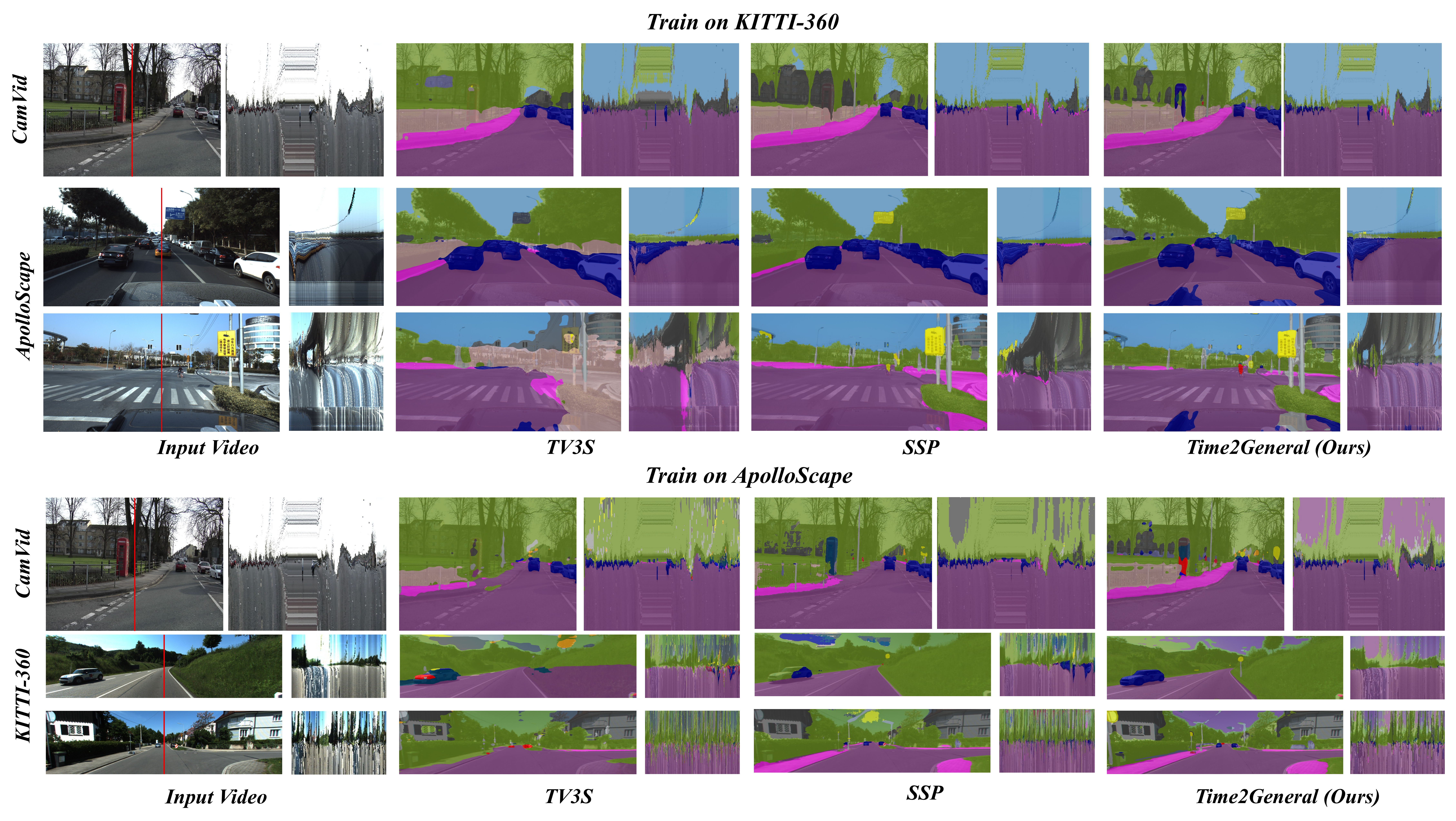}
  \setlength{\abovecaptionskip}{-0.1cm}
  \setlength{\belowcaptionskip}{-0.2cm}
  \caption{Qualitative comparison of \textit{KITTI-360 $\rightarrow$ CamVid, Apollo.} and \textit{Apollo. $\rightarrow$ CamVid, KITTI-360} predictions.}
  \label{fig: killresult}
\end{figure*}

\textbf{Results under three single-source training protocols.} Tables~\ref{tab:KITTI-360}--\ref{tab:CamVid} reports DGVSS results evaluated on multiple unseen targets. Time2General consistently achieves the best overall performance across all three protocols, improving both segmentation accuracy on mIoU and temporal consistency on mVC$_8$/mVC$_{16}$. Figure~\ref{fig: killresult} provides qualitative comparisons of training on KITTI-360 and ApolloScape. In all cases, Time2General produces cleaner boundaries and fewer broken regions. Across consecutive frames, Time2General better preserves object identity and label stability, leading to visibly reduced flicker. This observation aligns with the consistent gains in the quantitative assessment. 



Table~\ref{tab:spd_param_1024x512} demonstrates that Time2General achieves the highest throughput (18.15 FPS), substantially faster than the DGSS baselines (3.85--6.25 FPS), and also faster than the recent VSS-based SSP (10.99 FPS) and TV3S (6.15 FPS). This efficiency mainly comes from our lightweight query-based decoder on top of a frozen backbone, which avoids per-frame expensive costs while maintaining temporally consistent predictions.

\subsection{Ablation Studies}

\begin{table}[!t]
  \centering
  \caption{Comparison of inference speed at $1024\times512$ resolution.}
  \label{tab:spd_param_1024x512}
  \resizebox{0.75\linewidth}{!}{
  \begin{tabular}{lccc}
    \toprule
    Method & Backbone & \makecell{Speed\\(FPS)} & \makecell{Train Params\\(M)} \\
    \midrule
    REIN & DINOv2-L & 6.25 & 23.56 \\
    FADA & DINOv2-L & 3.93 & 22.64 \\
    DepthForge & DINOv2-L & 3.85 & 27.90 \\
    SSP & SAM-S & 10.99 & 61.62 \\
    TV3S & MiT-B5 & 6.15 & 85.48 \\
    \midrule
    Time2General (Ours) & DINOv2-L & 18.15 & 24.46 \\
    \bottomrule
  \end{tabular}
  }
\end{table}

\begin{table}[!t]
\caption{
Ablation on key components under KITTI-360 $\rightarrow$ Apollo. + CamVid. (\%)
}
\label{tab:ablation_A}
\centering
\resizebox{\linewidth}{!}{
\begin{tabular}{c | l | ccc | ccc}
\toprule
\multirow{2}{*}{\#} & \multirow{2}{*}{Variant}
& \multicolumn{3}{c|}{CamVid}
& \multicolumn{3}{c}{Apollo.} \\
\cmidrule(lr){3-5}
\cmidrule(lr){6-8}
&
& mIoU & mVC$_8$ & mVC$_{16}$
& mIoU & mVC$_8$ & mVC$_{16}$ \\
\midrule
1 & DepthForge (baseline)
& 51.28 & 84.65 & 84.26
& 43.87 & 82.90 & 84.11 \\

2 & + Stability Queries
& 60.12 & 89.10 & 88.92
& 46.02 & 84.95 & 85.60 \\

3 & + Spatio Temporal Memory Decoder
& 64.35 & 91.55 & 91.20
& 47.60 & 86.70 & 88.15 \\

4 & + Randomized training strides
& 58.90 & 88.20 & 88.05
& 45.50 & 84.10 & 85.10 \\

5 & + Masked Temporal Consistency Loss
& 66.93 & 93.76 & 93.53
& 48.63 & 91.07 & 92.73 \\

6 & Time2General
& \textbf{72.09} & \textbf{94.56} & \textbf{94.76}
& \textbf{50.51} & \textbf{91.73} & \textbf{94.76} \\
\bottomrule
\end{tabular}
}
\end{table}

\begin{table}[!t]
\caption{
Effect of Masked Temporal Consistency Loss on image-based DGSS baselines. MTC uses the same configuration as described in Implementation Details. (\%)
}
\label{tab:mtc_on_image}
\centering
\resizebox{\linewidth}{!}{
\begin{tabular}{c | l | ccc | ccc}
\toprule
\multirow{2}{*}{\#} & \multirow{2}{*}{Method}
& \multicolumn{3}{c|}{CamVid}
& \multicolumn{3}{c}{Apollo.} \\
\cmidrule(lr){3-5}
\cmidrule(lr){6-8}
& 
& mIoU & mVC$_8$ & mVC$_{16}$
& mIoU & mVC$_8$ & mVC$_{16}$ \\
\midrule
1 & REIN~\cite{wei2024stronger} 
& 41.19 & 84.03 & 83.53
& 38.92 & 81.84 & 83.36 \\

2 & REIN + MTC Loss
& 60.17 & 93.79 & 93.42
& 40.53 & 90.60 & 92.37 \\
\midrule

3 & FADA~\cite{bi2024learning} 
& 55.51 & 83.07 & 82.00
& 44.42 & 81.42 & 82.57 \\

4 & FADA + MTC Loss
& 67.15 & 93.00 & 91.77
& 48.77 & 90.31 & 91.53 \\
\midrule

5 & DepthForge~\cite{chen2025stronger}
& 51.28 & 84.65 & 84.26
& 43.87 & 82.90 & 84.11 \\

6 & DepthForge + MTC Loss
& 66.93 & 93.76 & 93.53
& 48.63 & 91.07 & 92.73 \\
\bottomrule
\end{tabular}
}
\end{table}

Table~\ref{tab:ablation_A} analyzes major design settings. We compare variants with and without Stability Queries under the same frozen depth prior, showing that the gains mainly stem from Stability Queries for multimodal alignment and temporal decoding rather than merely using the depth prior. Introducing the Spatio-Temporal Memory Decoder and Masked Temporal Consistency improves both accuracy and temporal stability, and the full model achieves the best mIoU of 72.09\%/50.51\% on CamVid/Apollo., with consistently the highest mVC metrics.

Table~\ref{tab:mtc_on_image} shows that adding MTC consistently improves temporal stability across all baselines, yielding substantial mVC gains on both targets. mIoU also increases, suggesting that temporal regularization can aid cross-domain generalization rather than merely smoothing predictions. We keep all other training settings unchanged, making MTC a simple plug-and-play objective for improving temporal coherence and robustness in DGSS.

\section{Conclusion}

In this paper, we present Time2General, a query-driven framework for domain-generalized video semantic segmentation. By freezing a VFM backbone and introducing a compact set of Stability Queries as temporally persistent semantic anchors, our method learns robust multi-scale representations and integrates complementary cues such as geometry and text with minimal risk of overfitting. Built on query-conditioned pixel features, a Spatio-Temporal Memory Decoder aggregates multi-frame context in a correspondence-free manner, while a Masked Temporal Consistency Loss further suppresses flicker under multi-stride sampling. Extensive experiments across five driving video benchmarks and corruption settings demonstrate improved cross-domain generalization and temporal stability, suggesting a practical path toward stable video segmentation under real-world domain and temporal errors.

\section*{Impact Statement}
This paper presents work whose goal is to advance the field of Machine Learning. There are many potential societal consequences of our work, none which we feel must be specifically highlighted here.


\bibliography{example_paper}
\bibliographystyle{icml2026}

\newpage
\appendix
\onecolumn
\section{More Related Works}

\subsection{Video Semantic Segmentation}

Video Semantic Segmentation (VSS) aims to assign pixel-level semantic labels to every frame in a video sequence \cite{ravi2024sam}. Unlike per-frame independent inference, VSS crucially benefits from exploiting inter-frame redundancy and motion cues to capture temporal dependencies, thereby improving per-frame accuracy while enhancing the temporal consistency of predictions. Existing approaches to temporal modeling can be broadly grouped into two paradigms: propagation-based methods and clip-level aggregation methods. 

Propagation-based methods leverage optical flow and learned motion fields to align and propagate semantic information from reference frames (e.g., keyframes and historical frames) to target frames via warping \cite{zhu2017deep,jain2019accel,gadde2017semantic,karaev2024cotracker}, followed by fusion with features from the current frame. 
Lightweight refinement modules are often introduced to further improve prediction quality \cite{liu2020efficient}. By reusing semantics from reference frames, propagation-based approaches can improve cross-frame stability and reduce redundant per-frame computation. More recently, semantic similarity based propagation has been explored to further improve temporal consistency in semi-supervised VSS \cite{vincent2025high}.

In contrast, clip-level methods take short consecutive clips as input and jointly aggregate multi-frame information within a clip using temporal attention \cite{wang2018non}, cross-frame affinity modeling \cite{miao2021vspw}, and memory-based aggregation modules \cite{wang2021temporal,cheng2022masked,shin2024video}. This design provides richer temporal context for predicting the current frame, improving accuracy and mitigating temporal flickering. Recent work also explores temporal state space sharing as an efficient mechanism for modeling temporal dependencies across frames \cite{hesham2025exploiting}. Beyond pure semantic labeling, universal video segmentation frameworks aim to unify multiple video segmentation tasks and further emphasize video-level consistency \cite{zhang2025dvis++}.

Despite substantial progress under in-domain settings, cross-domain generalization remains considerably more challenging. 
Propagation-based methods are particularly vulnerable to unreliable motion estimation under adverse weather, low illumination, and noisy conditions \cite{sakaridis2021acdc,yang2024depth}, which can induce alignment errors that accumulate and amplify over time. 
Clip-level methods often depend on fixed clip lengths and sampling intervals. When frame rates vary substantially and sampling becomes irregular, the gains from temporal modeling can become unstable.

\subsection{Video Generalizable Semantic Segmentation}

Video Generalizable Semantic Segmentation (VGSS) aims to jointly model temporal continuity in videos and cross-domain generalization, thereby alleviating performance degradation under distribution shift \cite{zhang2024video}. Prior studies are typically categorized into Unsupervised Domain Adaptation (UDA) and Domain Generalization (DG). 

UDA leverages unlabeled target domain data during training for distribution alignment and self-training \cite{shin2021unsupervised}, and it often incorporates temporal-consistency regularization to reduce cross-frame instability \cite{xing2022domain}. However, in realistic settings where the target domain is unavailable beforehand and changes dynamically, DG is more practical because it learns without relying on any target domain data \cite{bi2024learning}. 

Compared with domain generalization for image semantic segmentation (DGSS), VGSS remains less explored and has not yet converged to a unified framework. Consequently, many methods borrow strategies from DGSS and extend them to the video setting \cite{wei2024stronger}. Concretely, DGSS approaches mainly fall into two families: style augmentation/domain randomization and domain-invariant representation learning. The former synthesizes diverse “pseudo-domains” by perturbing color, texture, and frequency spectra, reducing over-reliance on the source domain \cite{lee2022wildnet}. The latter suppresses domain-specific factors and learns semantic representations via normalization and contrastive/consistency-based constraints. 

Extending domain generalization to videos introduces additional temporal challenges: domain shifts can disrupt cross-frame correspondences, making temporal inconsistencies more likely to emerge and accumulate over time \cite{sakaridis2021acdc}. 
To mitigate this issue, one line of VGSS work stabilizes predictions through cross-frame consistency constraints, such as class-wise statistics matching and prototype alignment \cite{huang2021cross}, while others combine style augmentation or domain-invariant learning with explicit temporal constraints to further enhance robustness. Nevertheless, domain shifts exacerbate difficulties caused by occlusions and scene changes, under which temporal dependencies are easily compromised \cite{karaev2024cotracker}. These observations underscore the need for resilient cross-domain temporal modeling strategies.

\section{Dataset and Label Alignment}\label{app: label}

The details of the evaluated datasets are as follows:

\textbf{KITTI-360} continuous driving trajectories with dense semantic annotations serve as our long-sequence domain.

\textbf{ApolloScape} is also collected along long driving traces with large-scale pixel-wise annotations, forming another long-sequence domain.

\textbf{CamVid} provides annotated driving videos and is treated as a short-sequence domain.

\textbf{Cityscapes-sequence} consists of fixed-length snippets (30 frames) and also serves as a short-sequence domain.

\textbf{Cityscapes-sequence-Corrupted} is used as an additional held-out test set, applying four weather corruptions (\textit{fog, frost, snow, spatter}) to the same Cityscapes snippets.

We align the label spaces across all datasets and evaluate on \textbf{15 shared classes}: \{\texttt{road}, \texttt{sidewalk}, \texttt{building}, \texttt{wall}, \texttt{fence}, \texttt{pole}, \texttt{traffic light}, \texttt{traffic sign}, \texttt{vegetation}, \texttt{sky}, \texttt{person}, \texttt{rider}, \texttt{car}, \texttt{truck \& bus}, \texttt{motorcycle}\}. A detailed label mapping from each dataset's original taxonomy to this 15-class space is provided in Table~\ref{tab:id_mapping_15_overlap}. For a given dataset, some classes may exist in its taxonomy but have zero ground-truth pixels in the evaluation split; we exclude such classes from the per-dataset mean and report average scores over valid classes only.

\begin{table}[h]
\centering
\small
\setlength{\tabcolsep}{6pt}
\renewcommand{\arraystretch}{1.15}
\caption{ID mapping tables for the 15 fully-overlapped classes across Cityscapes, ApolloScape, and CamVid.}
\label{tab:id_mapping_15_overlap}
\begin{tabular}{c l c c c}
\toprule
Overlap ID & Class & Cityscapes TrainID & ApolloScape TrainID & CamVid ID \\
\midrule
0  & road          & 0    & 9   & 17 \\
1  & sidewalk      & 1    & 10  & 19 \\
2  & building      & 2    & 20  & 4  \\
3  & wall          & 3    & 17  & 30 \\
4  & fence         & 4    & 13  & 9  \\
5  & pole          & 5    & 15  & 8  \\
6  & traffic light & 6    & 14  & 24 \\
7  & traffic sign  & 7    & 16  & 20 \\
8  & vegetation    & 8    & 21  & 26+29 \\
9  & sky           & 10   & 0   & 21 \\
10 & person        & 11   & 4   & 16 \\
11 & rider         & 12   & 5   & 2  \\
12 & car           & 13   & 1   & 5  \\
13 & truck\_bus    & 14+15 & 6+7 & 27 \\
14 & motorcycle    & 17   & 2   & 13 \\
\bottomrule
\end{tabular}

\vspace{2pt}
\footnotesize
\textit{Notes:} (1) vegetation: CamVid merges Tree (26) and VegetationMisc (29). 
(2) truck\_bus: Cityscapes merges truck (14) and bus (15); ApolloScape merges truck (6) and bus (7).
\end{table}

\section{Metrics}\label{app: metrics}
We use mean video consistency (mVC) with a window size $n$. Given a video $v$ with $C_v$ frames ($C_v\ge n$), let $M_t$ be the ground-truth mask on frame $t$, and $O_t$ be the predicted per-pixel labels. We compute consistency under dense-GT evaluation and sparse-GT (approx.) evaluation as follows:
\begin{equation}
\begin{aligned}
\mathrm{VC}^{\text{dense}}_n(v)
&=\frac{1}{C_v-n+1}\sum_{t=1}^{C_v-n+1}
\frac{\left|\bigcap_{i=t}^{t+n-1} M_i\ \cap\ \bigcap_{i=t}^{t+n-1} O_i\right|}
{\left|\bigcap_{i=t}^{t+n-1} M_i\right|},
\end{aligned}
\end{equation}
\begin{equation}
\begin{aligned}
\mathrm{VC}^{\text{approx}}_n(v)
&=\frac{1}{|\mathcal{R}_v|}\sum_{r\in\mathcal{R}_v}
\frac{\left|\,M_r\ \cap\ \bigcap_{i=r}^{r+n-1} S_{r,i}\ \cap\ \bigcap_{i=r}^{r+n-1} O_i\,\right|}
{\left|\,M_r\ \cap\ \bigcap_{i=r}^{r+n-1} S_{r,i}\,\right|}.
\end{aligned}
\end{equation}
where $M_i$ denotes label-stable ground-truth pixels within the window, and $S_{r,i}$ is a static-pixel mask with respect to the labeled reference frame $r$, obtained by grayscale-difference. We report $\mathrm{mVC}_n$ by averaging $\mathrm{VC}_n(v)$ over videos.

\section{Limitations and future work}

Although our model clearly outperforms the baselines in both accuracy and inference speed, it still faces challenges when handling streaming videos, which we will explore in future work.

\section{More Experiment}

Table \ref{tab:tau_sensitivity} and Figure \ref{fig:tau_placeholder} analyze the effect of the trimmed mean ratio $\tau$ on performance and optimization. In Table \ref{tab:tau_sensitivity}, $\tau{=}0.2$ yields the best overall results on both target domains, achieving the highest or near highest mIoU and mVC on CamVid and the best scores across all metrics on ApolloScape. Smaller trimming ratios underperform for different reasons: $\tau{=}0.0$ applies no trimming, so the MTC objective is more affected by outlier pixel pairs with large discrepancies, which can bias optimization and reduce robustness; $\tau{=}0.1$ partially suppresses outliers but remains insufficient, consistent with its persistently higher and noisier training loss in Figure \ref{fig:tau_placeholder}. Figure \ref{fig:tau_placeholder} further shows consistent optimization behavior: $\tau{=}0.2$ converges faster and to a lower MTC loss, whereas $\tau{=}0.3$ exhibits larger early fluctuations, likely because excessive trimming reduces the number of effective pixel pairs, increasing gradient variance and weakening supervision.

\begin{table}[h]
\centering

\begin{minipage}[c]{0.50\linewidth}
\centering
\captionof{table}{Sensitivity analysis of trimmed-mean ratio $\tau$.}
\label{tab:tau_sensitivity}
\begin{tabular}{c|ccc|ccc}
\toprule
\multirow{2}{*}{$\tau$}
& \multicolumn{3}{c|}{CamVid}
& \multicolumn{3}{c}{Apollo.} \\
\cmidrule(lr){2-4}\cmidrule(lr){5-7}
& mIoU & mVC$_8$ & mVC$_{16}$ & mIoU & mVC$_8$ & mVC$_{16}$ \\
\midrule
0.0 & 71.74 & 94.14 & 94.24
    & 50.48 & 91.41 & 92.89 \\
0.1 & 71.90 & 94.10 & 94.51
    & 49.32 & 91.51 & 93.13 \\
0.2 & \textbf{72.09} & \textbf{94.56} & \textbf{94.76}
    & \textbf{50.51} & \textbf{91.73} & \textbf{93.34} \\
0.3 & 71.62 & 94.34 &  94.52
    & 48.47 & 90.78 &  92.47\\
\bottomrule
\end{tabular}
\end{minipage}
\hfill
\begin{minipage}[c]{0.45\linewidth}
\centering
\includegraphics[width=\linewidth,height=0.22\textheight,keepaspectratio]{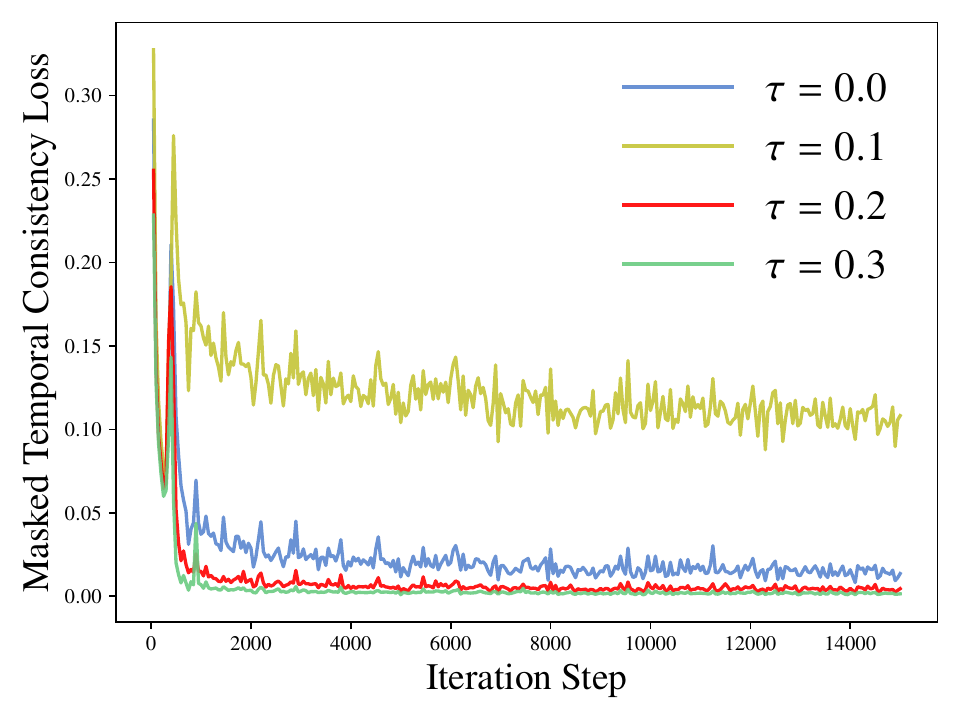}
\captionof{figure}{Effect of Trimming Ratio on MTC Loss.}
\label{fig:tau_placeholder}
\end{minipage}

\end{table}

\begin{figure*}[h]
  \centering
  \includegraphics[width=0.95\textwidth]{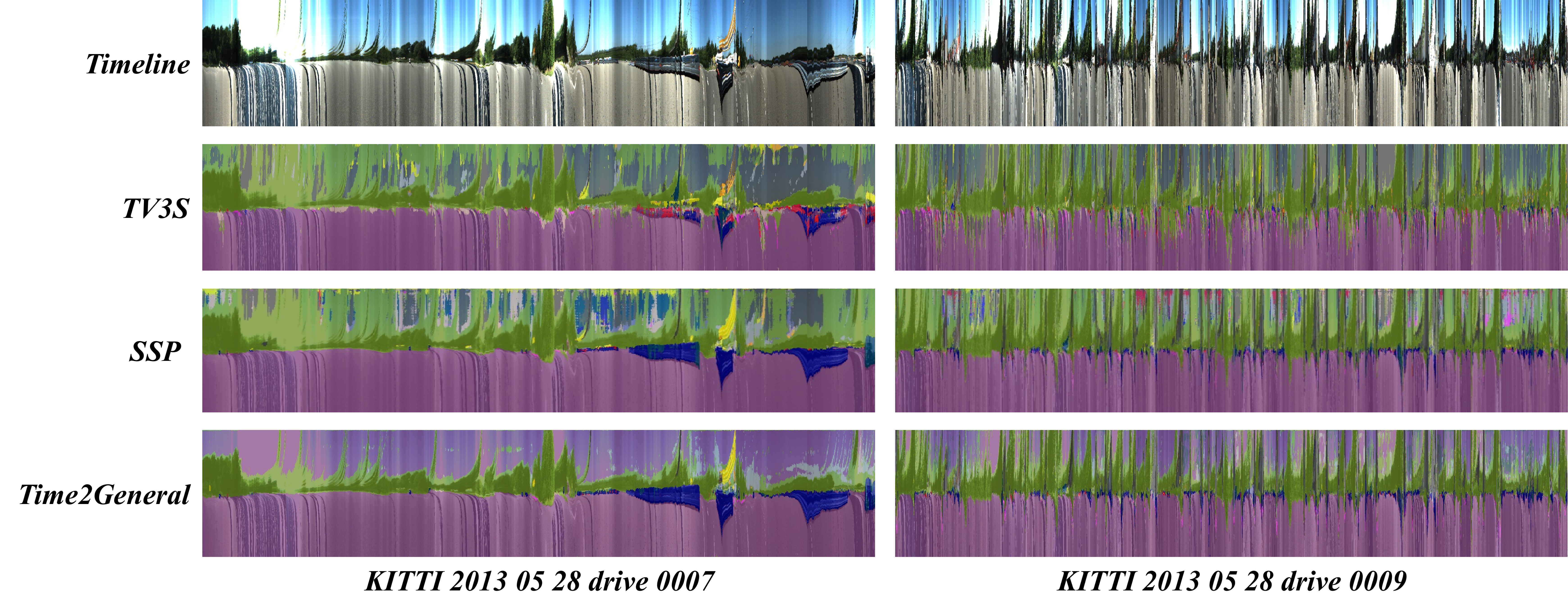}
  \caption{Qualitative comparison for real-world long video Semantic Segmentation under \textit{Apollo. $\rightarrow$ KITTI-360}.}
  \label{fig: supptimeline}
\end{figure*}

Figure \ref{fig: supptimeline} presents a qualitative comparison of real-world long-video semantic segmentation. We select two KITTI sequences as examples, with each column showing a timeline visualization of one video. For each method, the strip concatenates per-frame predictions along the time axis, enabling a clear inspection of temporal stability and class switching. Overall, other methods exhibit more fragmented patterns and frequent local label flips under cross-domain deployment, where the same object or static region often changes its predicted class and the boundaries drift over time. In contrast, Time2General yields a more coherent timeline, with smoother boundary evolution and more consistent labels in stable regions.

\begin{figure*}[h]
  \centering
  \includegraphics[width=0.95\textwidth]{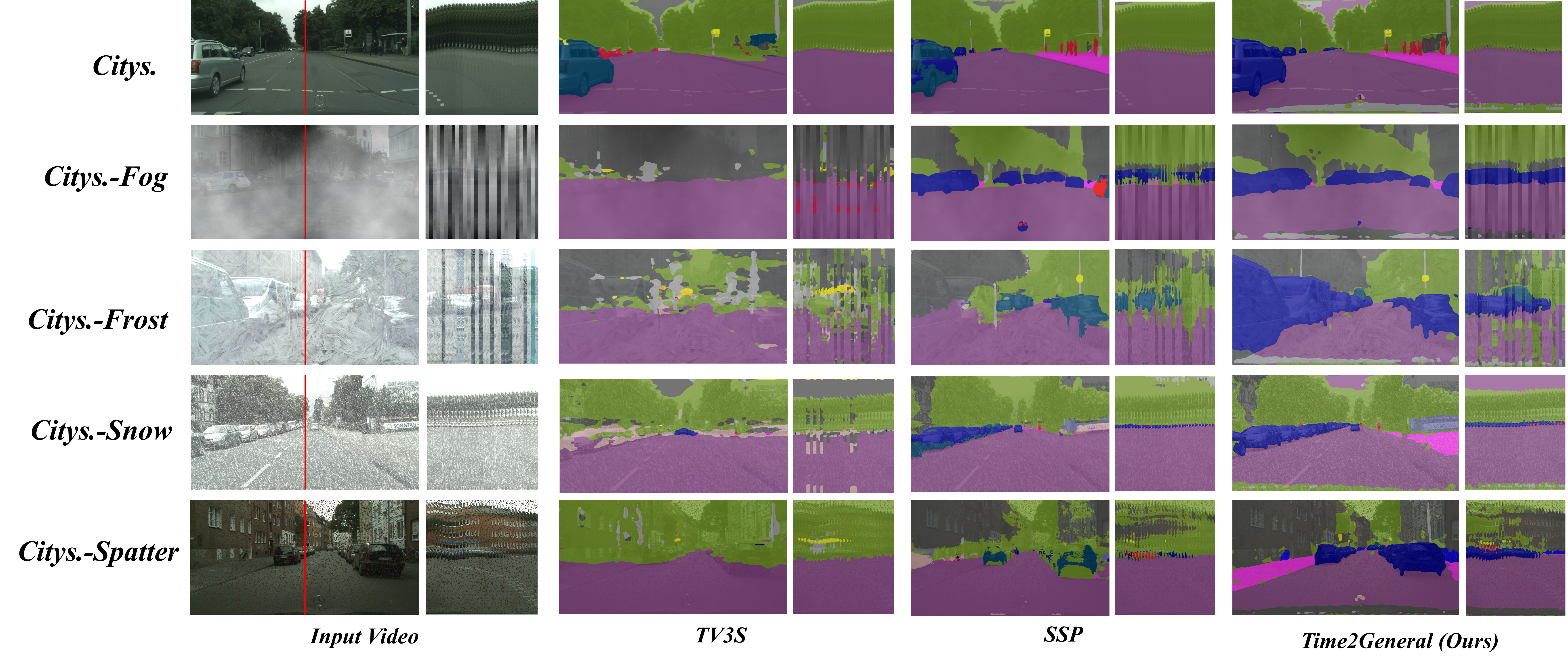}
  \caption{Qualitative comparison of cross domain segmentation predictions for \textit{Apollo. $\rightarrow$ City.-s + City.-s-C}.}
  \label{fig: killapollo2city}
\end{figure*}

Figure \ref{fig: killapollo2city} presents a qualitative comparison under the cross-domain setting \textit{Apollo. $\rightarrow$ Citys.-s + Citys.-s-C}. Overall, existing methods are more prone to local misclassification, class switching, and fragmented boundaries under fog, frost, snow, and spatter conditions, especially in distant regions and thin structures where predictions often become discontinuous and exhibit noticeable noisy streaks. In contrast, Time2General maintains a more coherent scene layout and more stable object boundaries across diverse weather domains, with higher label consistency in static regions, demonstrating stronger cross-domain robustness and temporal stability.

\end{document}